\documentclass{article}

\usepackage{makeidx}  
\usepackage{graphicx}
\usepackage{cite}

\usepackage{color}
\usepackage{amsmath}
\usepackage{epsfig}
\usepackage{url}

\usepackage{bm}

\newcommand{\A}{{\bf A}}
\newcommand{\B}{{\bf B}}

\newcommand{\I}{{\bf I}}

\newcommand{\x}{{\bf x}}
\newcommand{\X}{{\bf X}}

\newcommand{\y}{{\bf y}}

\title{Computing $p$-values of LiNGAM outputs via Multiscale Bootstrap}
\author{Yusuke Komatsu\thanks{Department of Mathematical and Computing Sciences, Tokyo Institute of Technology, 2-12-1 Ookayama, Meguro-ku, Tokyo 152-8552, Japan.}, Shohei Shimizu\thanks{The Institute of Scientific and Industrial Research (ISIR), Osaka University, Mihogaoka 8-1, Ibaraki, Osaka 567-0047, Japan. Email: sshimizu@ar.sanken.osaka-u.ac.jp}, Hidetoshi Shimodaira\thanks{Department of Mathematical and Computing Sciences, Tokyo Institute of Technology, 2-12-1 Ookayama, Meguro-ku, Tokyo 152-8552, Japan}
}
\date{}

\begin{document}

\maketitle

\begin{abstract}
Structural equation models and Bayesian networks have been widely used to study causal relationships between continuous variables.
Recently, a non-Gaussian method called LiNGAM was proposed to discover such causal models and has been extended in various directions. An important problem with LiNGAM is that the results are affected by the random sampling of the data as with any statistical method. Thus, some analysis of the confidence levels should be conducted. A common method to evaluate a confidence level is a bootstrap method. However, a confidence level computed by ordinary bootstrap is known to be biased as a probability-value ($p$-value) of hypothesis testing. In this paper, we propose a new procedure to apply an advanced bootstrap method called multiscale bootstrap to compute $p$-values of LiNGAM outputs. The multiscale bootstrap method gives unbiased $p$-values with asymptotic much higher accuracy. Experiments on artificial data demonstrate the utility of our approach.
 \end{abstract}

\section{Introduction}\label{sec:intro}
Structural equation models \cite{Bollen89book} and Bayesian networks \cite{Pearl00book,Spirtes93book} have been widely applied to analyze causal relationships in many fields. 
Many methods \cite{Pearl00book,Spirtes93book} have been developed to discover such a causal model when no prior knowledge on the network structure is available. 
Recently, a non-Gaussian method called LiNGAM \cite{Shimizu06JMLR} was proposed. 
The new method estimates a causal ordering of variables using passive observational data alone. 
The estimated ordering is correct if the causal relations form a linear structural equation model with non-Gaussian external influence variables {\it and} the sample size is {\it infinitely large}. 
In practice, however, the sample size is finite. 
The finite sample size induces statistical errors in the estimation, and the estimated ordering may not be right even when the model assumptions are reasonable. 
Thus, some analysis of the statistical reliability or confidence level of the estimated ordering should be done. 
In this paper, we discuss such reliability analysis of LiNGAM. 

A common procedure to evaluate such a confidence level is statistical hypothesis testing \cite{Lehmann08Test}. 
In statistical testing, one computes a probability-value ($p$-value) of a hypothesis. The hypothesis is rejected when the $p$-value is  not greater than a pre-specified level of significance, say 5\%. 
There are several approaches to define a $p$-value. 
Bootstrapping \cite{Efron93book} is a well-known computational method for computing confidence levels when a simple mathematical formula is difficult to derive. 
It is a resampling method to approximate a random sample by a bootstrap sample that is created by random sampling with replacement from the original single dataset. 
Felsenstein \cite{Felsenstein85Evol} proposed to use bootstrapping to define a $p$-value in the context of phylogenetic tree selection of molecular evolution in bioinformatics. 
He defined a $p$-value of a tree by a frequency called bootstrap probability that the tree is found to be optimal when tree selection is performed for a number of bootstrap replicates of the original dataset. 
The idea has been applied to other multivariate analyses including Bayesian networks  \cite{Fried99UAI}. 

However, it is known that the bootstrap probability is biased as a $p$-value \cite{Efron96boot,Hillis93SB}. 
The naive bootstrapping tends to give overconfidence in wrong hypotheses. 
Thus, some advanced bootstrap methods to achieve higher accuracy have been proposed \cite{Efron96boot,Hall92book,Shimo02SB,Shimo08JSPI}. 
Among others, multiscale bootstrapping \cite{Shimo02SB,Shimo08JSPI} is much more accurate but still easy to implement and has been successful in the field of phylogenetic tree selection. 

In this paper, we propose to apply the multiscale bootstrap to compute confidence levels, {\it i.e.}, $p$-values, of variable orderings estimated by LiNGAM. 
The paper is structured as follows. 
First, in Section~\ref{sec:background}, we briefly review LiNGAM and multiscale bootstrap. 
In Section~\ref{sec:MB-lingam} we propose a new procedure to compute $p$-values of LiNGAM outputs using the multiscale bootstrap method. 
The multiscale bootstrap method is tested using artificial data in Section~\ref{sec:simulations}. 
Conclusions are given in Section~\ref{sec:conc}.

\section{Background}\label{sec:background}

\subsection{LiNGAM}\label{sec:lingam}

In \cite{Shimizu06JMLR}, a non-Gaussian variant of structural equation models and Bayesian networks, which is called LiNGAM, was proposed. 
Assume that observed data are generated from a process represented graphically by a directed acyclic graph, {\it i.e.}, DAG. 
Let us represent this DAG by a $m$$\times$$m$  adjacency matrix $\B$$=$$\{b_{ij}\}$ where every $b_{ij}$ represents the connection strength from a variable $x_j$ to another $x_i$ in the DAG, {\it i.e.}, the {\it direct} causal effect of $x_j$ on $x_i$. 
Let us further define $\A=(\I-\B)^{-1}$. The $(j,i)$-element $a_{ji}$ represents the {\it total} causal effect of $x_i$ on $x_j$ \cite{Hoyer07IJAR}.    
Moreover, let us denote by $k(i)$ a causal order of variables $x_i$ in the DAG so that no later variable influences any earlier variable. 
For example, a variable $x_j$ is not causally influenced by a variable $x_i$, {\it i.e.}, $a_{ji}$$=$$0$, if $k(j)<k(i)$.
Moreover, assume that the relations between variables are linear. 
Then we have
\begin{equation}
x_i = \sum_{k(j)<k(i)} b_{ij}x_j + e_i,\label{eq:model0}
\end{equation}
where $e_i$ is an external influence variable. 
All external influences $e_i$ are continuous random variables having \emph{non-Gaussian} distributions with zero means and non-zero variances, and $e_i$ are independent of each other so that there is no unobserved confounding variables \cite{Spirtes93book}. 
We emphasize that $k(j)$$<$$k(i)$ does not necessarily imply that $x_j$ influences $x_i$. 
It only implies that $a_{ji}$$=$$0$, and $a_{ij}$ can be either zero or non-zero. 
The causal ordering $k(i)$ only defines a {\it partial} order of variables, which is enough to define a DAG. 
In \cite{Shimizu06JMLR}, the LiNGAM model (\ref{eq:model0}) was shown to be identifiable without using any prior knowledge on the network structure. 
That is, the variable orders $k(i)$ and connection strengths $b_{ij}$ are estimable solely based on the data matrix of $\x=[x_1,\cdots,x_m]^T$ . 
In \cite{Shimizu06JMLR}, a discovery algorithm based on independent component analysis (ICA) \cite{Hyva01book}, which is called LiNGAM algorithm, was also proposed to estimate $k(i)$ and $b_{ij}$. 

\subsection{Bootstrap probability}
Denote by $\x$ a $m$-dimensional random variable vector and by $\X$$=$$(\x_1,\cdots,\x_n)$ a random sample of size $n$ from the distribution of $\x$. 
Further, define a function $f(\X)$ so that $f(\X)$$=$$0$ if a hypothesis is rejected and otherwise $f(\X)$$=$$1$. 
Suppose that we obtain a $m$$\times$$n$ data matrix $\overline{\X}$ that is generated from $\x$, and the function $f(\overline{\X})$$=$$1$. 
Then, it is useful to evaluate how statistically reliable the value of $f(\overline{\X})$$=$$1$ is since the function could return 0 for another data matrix due to sample fluctuation. 
In \cite{Felsenstein85Evol}, Felesenstein proposed to use bootstrapping \cite{Efron93book} to evaluate such reliability. 
Let us denote by $\X_q^{\ast}$ a $q$-th bootstrap sample of size $n^{\ast}$, which is created by random sampling with replacement from the columns of $\X$. 
In ordinary bootstrap, $n^{\ast}$ is taken to be $n$. 
Then, the bootstrap probability $p^{BP}$ is defined as a frequency that $f(\X^{\ast})$$=$1: 
\begin{eqnarray}
p^{BP} &=& \frac{1}{Q} \sum_{q=1}^{Q} f\left(\X_q^{\ast}\right),\label{eq:bp}
\end{eqnarray}
where $Q$ is the number of bootstrap replications. 
A testing procedure was proposed that the hypothesis is rejected if $p^{BP}$ is not greater than a significance level $\alpha$ ($0$$<$$\alpha$$<$$1$), say 0.05. 
However, it is known that $p^{BP}$ is {\it biased} as a $p$-value \cite{Efron96boot,Hillis93SB}. The multiscale bootstrap \cite{Shimo02SB,Shimo08JSPI} corrects the bias and gives a more accurate $p$-value. This is explained in more detail in the next subsection. 

\subsection{Unbiasedness}\label{sec:ub}
To discuss the bias of a $p$-value, it is conventional \cite{Efron96boot} to assume that 
there {\it exists} a function $g$ that transforms a random sample $\X$ to a $K$-dimensional random vector $\y$$=$$[y_1,\cdots,y_K]^T$ that (at least approximately) follows a Gaussian distribution with an unknown mean vector ${\bm \mu}$ and covariance identity $\I$, {\it i.e.}, $N_K({\bm \mu},\I)$.  
Note that it is {\it not} necessary to specify the actual functional form of $g$ and dimension $K$. 
Let us denote by $\mathcal{H}$ such a class of $\y$ that $f(\X)$$=$$1$.  
Then, the null hypothesis $f(\X)$$=$$1$ can be described as ${\bm \mu}$$\in$$\mathcal{H}$ in terms of a region in the parameter space. 
We only have to consider $\y$ to discuss the bias of a $p$-value computed based on $\X$ due to the {\it  transformation-respecting property} of bootstrapping \cite{Efron93book}.

In statistical hypothesis testing, the null hypothesis ${\bm \mu}$$\in$${\mathcal H}$ is rejected when a $p$-value computed based on $\y$, which is denoted by $p(\y)$, is not greater than a significance level $\alpha$. 
A test controls a type-I error if the probability of false rejection under the null hypothesis is not greater than $\alpha$. 
This is a desirable property of a testing procedure. 
Another desirable property is {\it unbiasedness} \cite{Lehmann08Test}.  
A test is unbiased if the  probability of correct rejection under alternative hypotheses is not less than $\alpha$, and the type-I error is also controlled. 
Then an unbiased test is formally defined to be a test that uses a $p$-value $p(\y)$ satisfying
\begin{eqnarray}
{\rm Prob}\{p(\y)<\alpha\}\le \alpha, \hspace{1mm} {\bm \mu} \in \mathcal{H}\hspace{3mm}{\rm and}\hspace{3mm}{\rm Prob}\{p(\y)<\alpha\}\ge \alpha, \hspace{1mm} {\bm \mu} \notin \mathcal{H}.
\end{eqnarray}
Let us denote by $\partial \mathcal{H}$ the boundary of $\mathcal{H}$. 
To satisfy the inequalities above, the following equation needs to hold \cite{Lehmann08Test}:
\begin{eqnarray}
{\rm Prob}\{p(\y)<\alpha\} = \alpha, \hspace{1mm} {\bm \mu} \in \partial \mathcal{H}.\label{eq:5}
\end{eqnarray}
In other words, $p(\y)$ follows a {\it uniform} distribution over the interval $[0,1]$. 
It has been shown \cite{Shimo02SB} that $p^{BP}$ has a rather large bias to meet the unbiasedness condition (\ref{eq:5}):
\begin{eqnarray}
{\rm Prob}\{p^{BP}(\y)<\alpha\} = \alpha + O(n^{-1/2}), 
\end{eqnarray}
where $O(\cdot)$ is the Landau symbol. 
Multiscale bootstrap \cite{Shimo02SB} reduces the bias. 
Let $p^{MB}$ denote a $p$-value computed by multiscale bootstrap. 
It can be shown that  $p^{MB}$ is approximately unbiased with asymptotic third-order accuracy:
\begin{eqnarray}
{\rm Prob}\{p^{MB}(\y)<\alpha\} = \alpha + O(n^{-3/2}),\label{eq:mb2}
\end{eqnarray}
Thus, multiscale bootstrap gives a $p$-value with much higher-order accuracy than ordinary bootstrap. 
Rigorously speaking, the boundary $\partial \mathcal{H}$ needs to be assumed to be smooth or approximately smooth. 
Otherwise, no unbiased test can be defined \cite{Lehmann08Test}. 
However, it has been shown that $p^{MB}$ is less biased than $p^{BP}$ even if the boundary is non-smooth \cite{Shimo08JSPI}.

\subsection{Multiscale Bootstrap}\label{sec:MB}
In \cite{Shimo08JSPI}, the theory of multiscale bootstrap \cite{Shimo02SB} was extended, and a class of unbiased $p$-values including $p^{MB}$ in (\ref{eq:mb2}) was obtained. 
Let $\y^{\ast}$ denote the $\y$ vector corresponding to $\X^{\ast}$$=$$[\x_1^{\ast}, \cdots, \x_{n^{\ast}}^{\ast}]$. Then
the standard deviation of $\y^{\ast}$ is proportional to $1/\sqrt{n^{\ast}}$, and its value
relative to the case $n^{\ast}$$=$$n$ is called `scale' of bootstrap resampling; this is
defined by $\sigma$$=$$\sqrt{n/n^{\ast}}$.
Then the bootstrap probability $p^{BP}$ in (\ref{eq:bp}) is a function of $\sigma^2$, which is denoted by  $p^{BP}_{\sigma^2}$ for clarity. 
The fundamental idea of the extended multiscale bootstrap \cite{Shimo08JSPI} is to compute the bootstrap probability $p^{BP}_{\sigma^2}$ with the scale $\sigma^2$$=$$-1$, {\it i.e.}, the bootstrap sample size $n^{\ast}$$=$$-n$. 
Of course, it is impossible to set $n^{\ast}$$=$$-n$. 
Therefore, one first select several scales $\sigma$$>$$0$, computes the bootstrap probability for each of the corresponding bootstrap sample sizes $n/\sigma^2$ and extrapolates the bootstrap probabilities to  $\sigma^2$$=$$-1$, {\it i.e.}, $n^{\ast}$$=$$-n$. 

We now review a procedure to compute such unbiased $p$-values. 
Let us define a bootstrap $z$-value by 
\begin{eqnarray}
z_{\sigma^2}=-\Phi^{-1}\left(p^{BP}_{\sigma^2}\right),\label{eq:nz}
\end{eqnarray}
where $\Phi^{-1}$ is the inverse of the distribution function $\Phi$ of the standard Gaussian distribution $N(0,1)$. 
Further, let us call $\sigma z_{\sigma^2}$ a normalized bootstrap $z$-value. 
Then, consider to model the changes in $\sigma z_{\sigma^2}$ along the changing the scale $\sigma$ by a model $\psi(\sigma^2|{\bm \beta})$, where ${\bm \beta}$$=$$[\beta_0, \cdots, \beta_{h-1}]^T$ is a parameter vector of the model. 
Two model classes are proposed in \cite{Shimo08JSPI}:
\begin{eqnarray}
\psi_1^{h}(\sigma^2|{\bm \beta})  &=& \sum_{j=0}^{h-1} \beta_j \sigma^{2j}, \hspace{1mm} h\ge1. \label{eq:ploy}\\
\psi_2^{h}(\sigma^2|{\bm \beta})  &=& \beta_0 + \sum_{j=1}^{h-2} \frac{\beta_j\sigma^{2j}}{1+\beta_{h-1}(\sigma-1)}, \hspace{1mm}  0\le \beta_{h-1}\le1,\hspace{1mm}  h\ge3.\label{eq:sing}
\end{eqnarray}
The model (\ref{eq:ploy}) is reasonable when the boundary $\partial \mathcal{H}$ is smooth, and the model (\ref{eq:sing}) is preferable when $\mathcal{H}$ is a cone and $\partial \mathcal{H}$ is not smooth. 
To estimate the models, a number of sets of bootstrap replicates with different scales $\sigma_d$ ($d$$=$$1$, $\cdots$, $D$) are first created, and subsequently the bootstrap probability $p^{BP}_{\sigma_d^2}$ for each scale is computed. 
Note that the bootstrap sample sizes may be different from that of the original dataset. 
Then, a set of scales and normalized bootstrap $z$-values \{$\sigma_d$, $\sigma_d z_{\sigma_d^2}$\} is obtained. 
Note that $z_{\sigma_d^2}$ is computed based on $p^{BP}_{\sigma_d^2}$ using (\ref{eq:nz}). 
Finally, the model parameter vector ${\bm \beta}$ are estimated using the set of scales and normalized bootstrap $z$-values. 
The maximum likelihood method is applied since $Qp^{BP}_{\sigma^2}$ follows a binomial distribution. A best model  $\psi_{best}(\sigma^2|{\bm \beta})$ is selected using an information criterion AIC \cite{Akaike74AIC}.

Then, a class of $p$-values using the best model $\psi_{best}(\sigma^2|{\bm \beta})$ is derived:
\begin{eqnarray}
p^{MB}_h &=& \Phi\left\{-\sum_{j=0}^{h-1}\frac{(-1-\sigma^2_0)^j}{j!}\frac{\partial^j \psi_{best}(\sigma^2|\hat{{\bm \beta}})}{\partial (\sigma^2)^j}\left|_{\sigma_0^2}\right.\right\},\label{eq:mbk}
\end{eqnarray}
where $\sigma_0^2$ is taken to be unity. 
The right side of (\ref{eq:mbk}) is the first $h$ terms of the Taylor series of the slope of $z_{\sigma^2}$ at $1/\sigma=1$, {\it i.e.}, $\partial z_\sigma^2/\partial(1/\sigma)|_1$, around $\sigma_0^2$. 
It can be shown that $p^{MB}_2$ is actually equal to $p^{MB}$ in (\ref{eq:mb2}) that achieves the unbiasedness with asymptotic third-order accuracy. 
Further, $p^{MB}_1$ turns out to be the naive bootstrap probability $p^{BP}$ in (\ref{eq:bp}). 
The larger $h$ gives an unbiased $p$-value with asymptotic higher-order accuracy \cite{Shimo08JSPI}.
However, it also makes the maximum likelihood estimation less stable. 
In practice, $h$$=$$2$ or $3$ is often used. 

\begin{figure}[h]
\begin{center}
\includegraphics[width=2in]{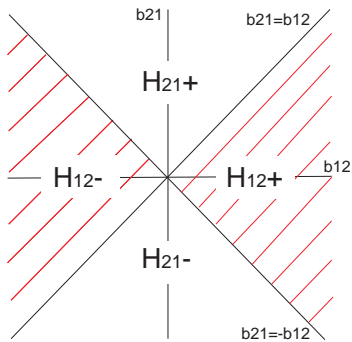}
\end{center}
\caption{Four regions $H_{12}^{+}$, $H_{12}^{-}$, $H_{21}^{+}$, and $H_{21}^{-}$.}
\label{fig:region}
\end{figure}

\section{A multiscale bootstrap procedure to assessing reliability of LiNGAM}\label{sec:MB-lingam}
We first define null hypotheses tested. 
We here focus on the following four types of hypotheses between $x_i$ and $x_j$ ($i$$\neq$$j$), although we can test hypotheses that describe the relations between more than two variables similarly: 
\begin{enumerate}
\item $H_{ij}^{+}$: a hypothesis that $x_i$ is directly caused by $x_j$, and its connection strength is positive, {\it i.e.}, $b_{ij}$$>$$0$; 
\item $H_{ij}^{-}$: a hypothesis that $x_i$ is directly caused by $x_j$, and its connection strength is negative, {\it i.e.}, $b_{ij}$$<$$0$; 
\item $H_{ji}^{+}$: a hypothesis that $x_j$ is directly caused by $x_i$, and its connection strength is positive, {\it i.e.}, $b_{ji}$$>$$0$; 
\item $H_{ji}^{-}$: a hypothesis that $x_j$ is directly caused by $x_i$, and its connection strength is negative, {\it i.e.}, $b_{ji}$$<$$0$. 
\end{enumerate}
See Fig.~\ref{fig:region} for the four regions of the parameter space in {\it two} variable cases {\it around the origin} that the connection strengths $b_{12}$ and $b_{21}$ are zeros. 
LiNGAM outputs $k(2)$$>$$k(1)$ if $|b_{12}|$$<$ $|b_{21}|$, and otherwise $k(1)$$>$$k(2)$ since each total effect $a_{ij}$ is equal to the corresponding direct effect $b_{ij}$ in two variable cases \cite{Shimizu06JMLR}.  
We note that the signs of connections strengths are important and interesting in many applications  \cite{Bollen89book,Silva06JMLR} as well as the variable orderings.
Further, this way of dividing the space based on the signs and orderings would make the boundaries of the regions be closer to be smooth than solely based on the orderings and help the multiscale bootstrap work better. 

We now propose a new procedure to apply Multiscale Bootstrap to LiNGAM, which we call {\it MB-LiNGAM}:

\noindent
  \rule{\columnwidth}{0.5mm}
       { \sffamily
	MB-LiNGAM procedure
	 \begin{enumerate}
	 \item \label{step:ica} Select the scales $\sigma_1$, $\cdots$, $\sigma_D$ ($D$$\ge$$2$) so that $n^{\ast}_d$$=$$n/\sigma_d^2$ is an integer and choose the number of bootstrap replicates $Q$. 
	 \item Generate $Q$ bootstrap replicates $\X^{\ast}_{q,d}$ ($q$$=$$1$, $\cdots$, $Q$) for each scale $\sigma_d$, {\it i.e.}, each bootstrap sample size $n^{\ast}_d$$=$$n/\sigma_d^2$.
	 \item\label{step:lingam} Perform LiNGAM algorithm to each bootstrap replicate $\X^{\ast}_{q,d}$ and compute the bootstrap probabilities $p^{BP}_d(H_{ij}^{+})$ and $p^{BP}_d(H_{ij}^{-})$ ($i$$\neq$$j$) for each scale $\sigma_d$, where $p^{BP}_d(H)$ denotes the bootstrap probability of a hypothesis $H$ for scale $\sigma_d$. 
	 \item Compute the multiscale bootstrap $p$-values $p^{MB}_h(H_{ij}^{+})$ and $p^{MB}_h(H_{ij}^{-})$ ($i$$\neq$$j$) using the procedure in Section \ref{sec:MB}, more specifically (\ref{eq:mbk}),  where $p^{MB}_h(H)$ denotes the multiscale bootstrap $p$-value of $H$ with the order $h$. 
	 \end{enumerate}
       } \vspace{-4mm}
\noindent \rule{\columnwidth}{0.5mm}
In the simulations below, the ICA part of LiNGAM algorithm is run several times in Step~\ref{step:lingam}. 
Each time the initial point of the optimization is randomly changed. 
The set of the estimates that achieves the largest value of an ICA objective function is used in the subsequent steps. 
It is a common practice to alleviate the effects of possible local maxima.


\paragraph{Related work}
Some methods have been proposed to test significance of direct effects $b_{ij}$\cite{Shimizu06JMLR,Hoyer07IJAR}.  
For simplicity, let us consider two variable cases, where each direct effect $b_{ij}$ is equal to the corresponding total effect $a_{ij}$ as mentioned above.
Those methods test if each of effects $b_{ij}$ is zero or not and imply that $k(i)$$<$$k(j)$ if `$b_{ij}$$=$$0$' is accepted and that $k(j)$$<$$k(i)$ if `$b_{ji}$$=$$0$' is accepted. Such a procedure would work if $b_{ij}$ or $b_{ji}$ is exactly zero. 
However, in reality, the assumptions of the model (\ref{eq:model0}) are more or less violated, and hence both of $b_{ij}$ and $b_{ji}$ could be non-zero. In such cases, those existing methods might reject both of the hypotheses and not give much information on which ordering is better. Even in the cases, our approach tells which ordering is better or statistically more reliable comparing bootstrap probabilities of the orderings. 

\section{Simulations}\label{sec:simulations}
We first created three LiNGAM models with $m$$=$$2$ variables: 
\begin{eqnarray}
\left[
\begin{array}{c}
x_1\\
x_2
\end{array}
\right]
& =& 
\left[
\begin{array}{cc}
0 & b\\
b & 0
\end{array}
\right]
\left[
\begin{array}{c}
x_1\\
x_2
\end{array}
\right]
+
\left[
\begin{array}{c}
e_1\\
e_2
\end{array}
\right],
\end{eqnarray}
where $b$$=$$0$, $0.01$ or $0.1$, and $e_1$ and $e_2$ followed a Laplace distribution with mean zero and variance two. 
The model with $b$$=$$0$ is on the boundary of $H_{12}^{+}$, $H_{12}^{-}$, $H_{21}^{+}$, and $H_{21}^{-}$. The model with $b$$=$$0.01$ or $0.1$ is on the boundary between $H_{12}^{+}$ and $H_{21}^{+}$. Further, we created two LiNGAM models with $m$$=$$6$ variables: 
\begin{eqnarray}
\left[
\begin{array}{c}
x_1\\
x_2\\
x_3\\
x_4\\
x_5\\
x_6
\end{array}
\right]
& =& 
\left[
\begin{array}{cccccc}
0 & 0 & 0 & 0 & 0 & 0\\
b & 0 & 0 & 0 & 0 & 0\\
b & 0 & 0 & 0 & 0 & 0\\
b & b & 0 & 0 & 0 & 0\\
0 & b & 0 & b & 0 & 0\\
b & b & b & 0 & b & 0
\end{array}
\right]
\left[
\begin{array}{c}
x_1\\
x_2\\
x_3\\
x_4\\
x_5\\
x_6
\end{array}
\right]
+
\left[
\begin{array}{c}
e_1\\
e_2\\
e_3\\
e_4\\
e_5\\
e_6
\end{array}
\right],
\end{eqnarray}
where $b$$=$$0$ or $0.5$, and $e_1$ and $e_2$ also followed a Laplace distribution whose mean zero and variance two. 
We randomly generated 1280 datasets with sample size 1000 under each of the five models. 
Then we applied MB-LiNGAM procedure in Section~\ref{sec:MB-lingam} to the datasets. 
The scales $\sigma_d$ were selected so that they gave integer values of bootstrap sample size and were (approximately) equally-spaced in log-scale between $1/9$ and $9$ ($d$$=$$1$, $\cdots$, $13$). 
The number of bootstrap replicates $Q$ was 1000, and the value $h$ for $p_h^{MB}$ was 3.

\begin{figure}[!b]
\centering
\begin{minipage}{.25\textwidth}
\centering
BP, $b=0$

\includegraphics[width=1.3in]{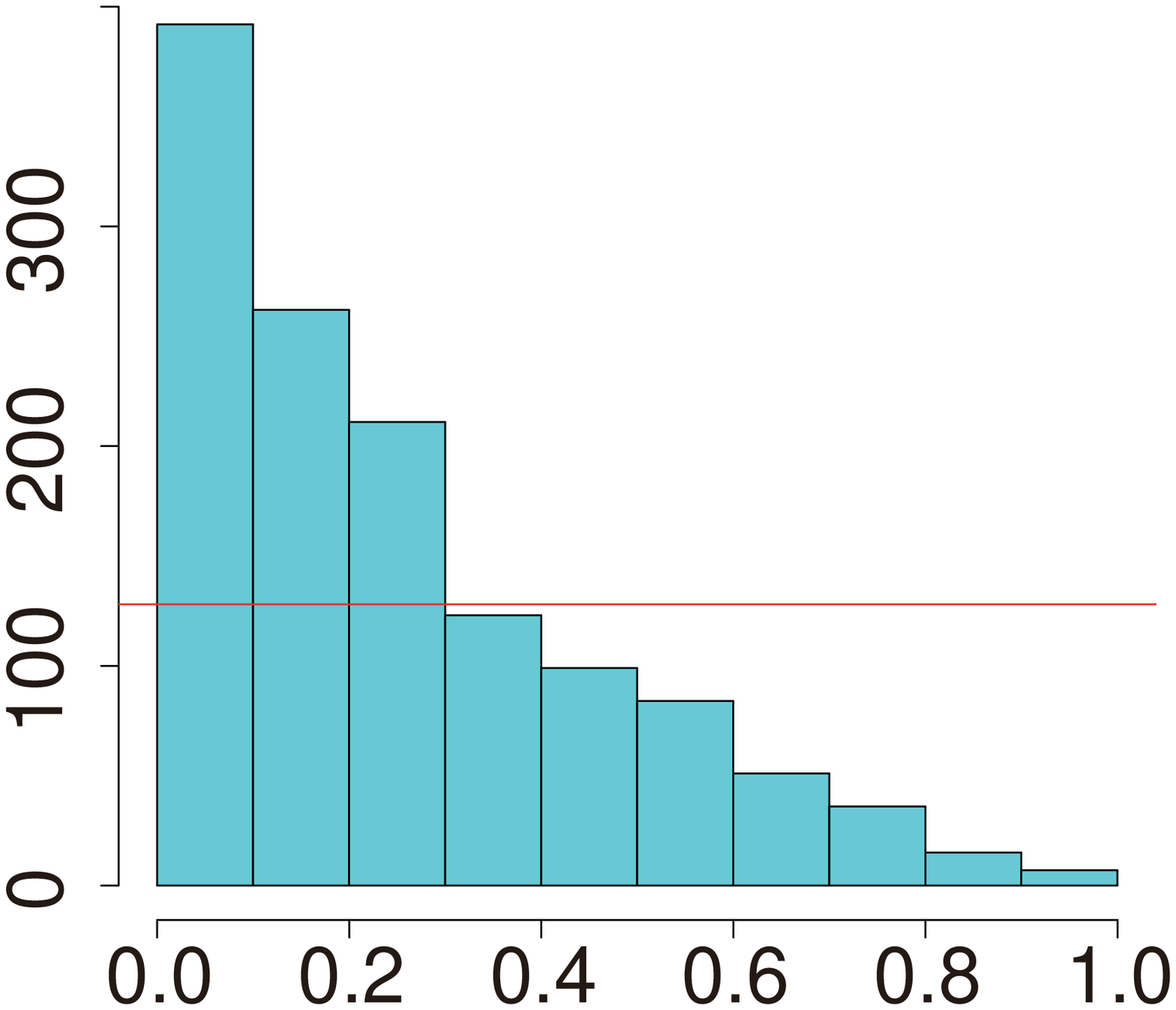}
\end{minipage}
\begin{minipage}{.25\textwidth}
\centering
BP, $b=0.01$

\includegraphics[width=1.3in]{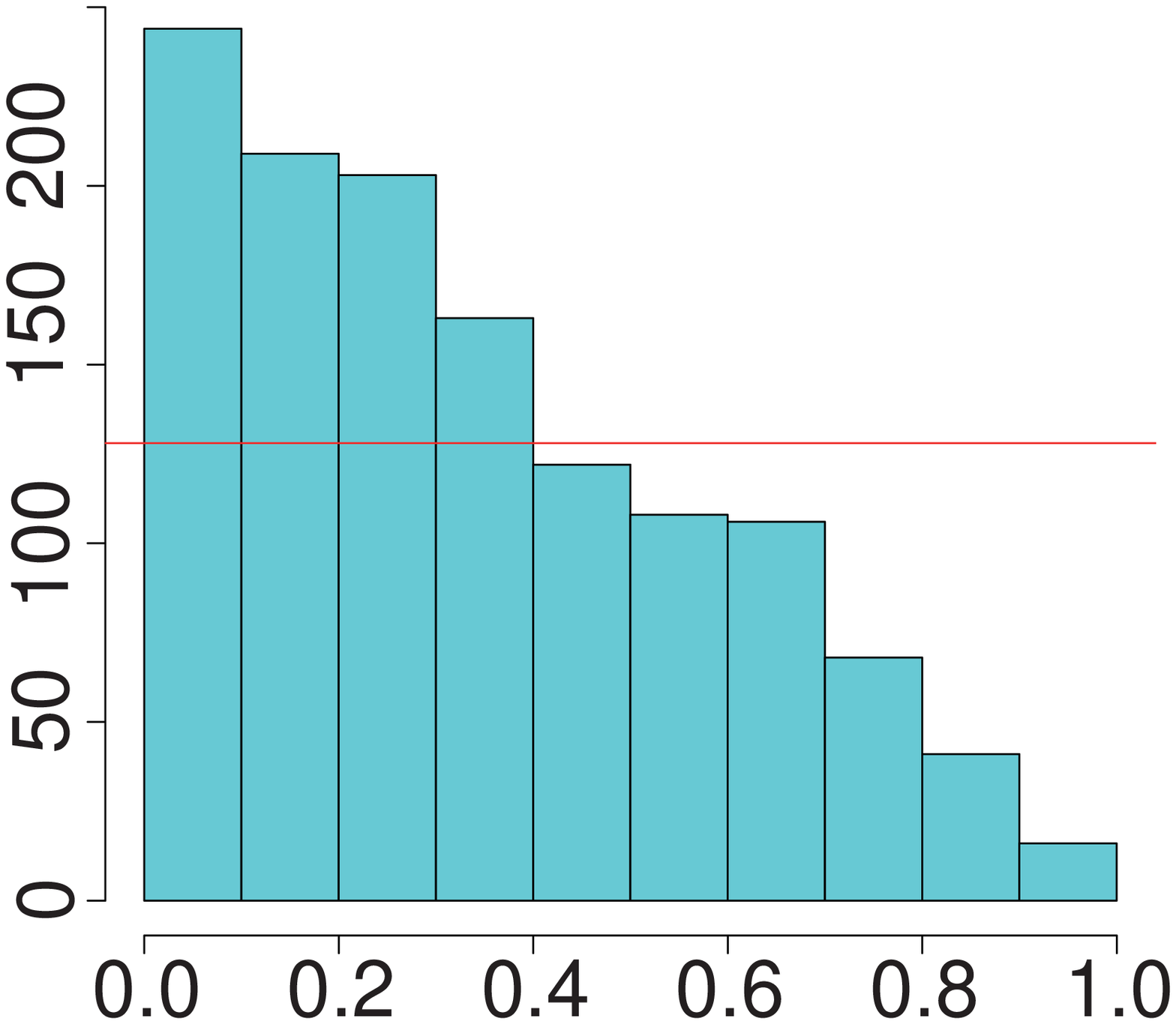}
\end{minipage}
\begin{minipage}{.25\textwidth}
\centering
BP, $b=0.1$

\includegraphics[width=1.3in]{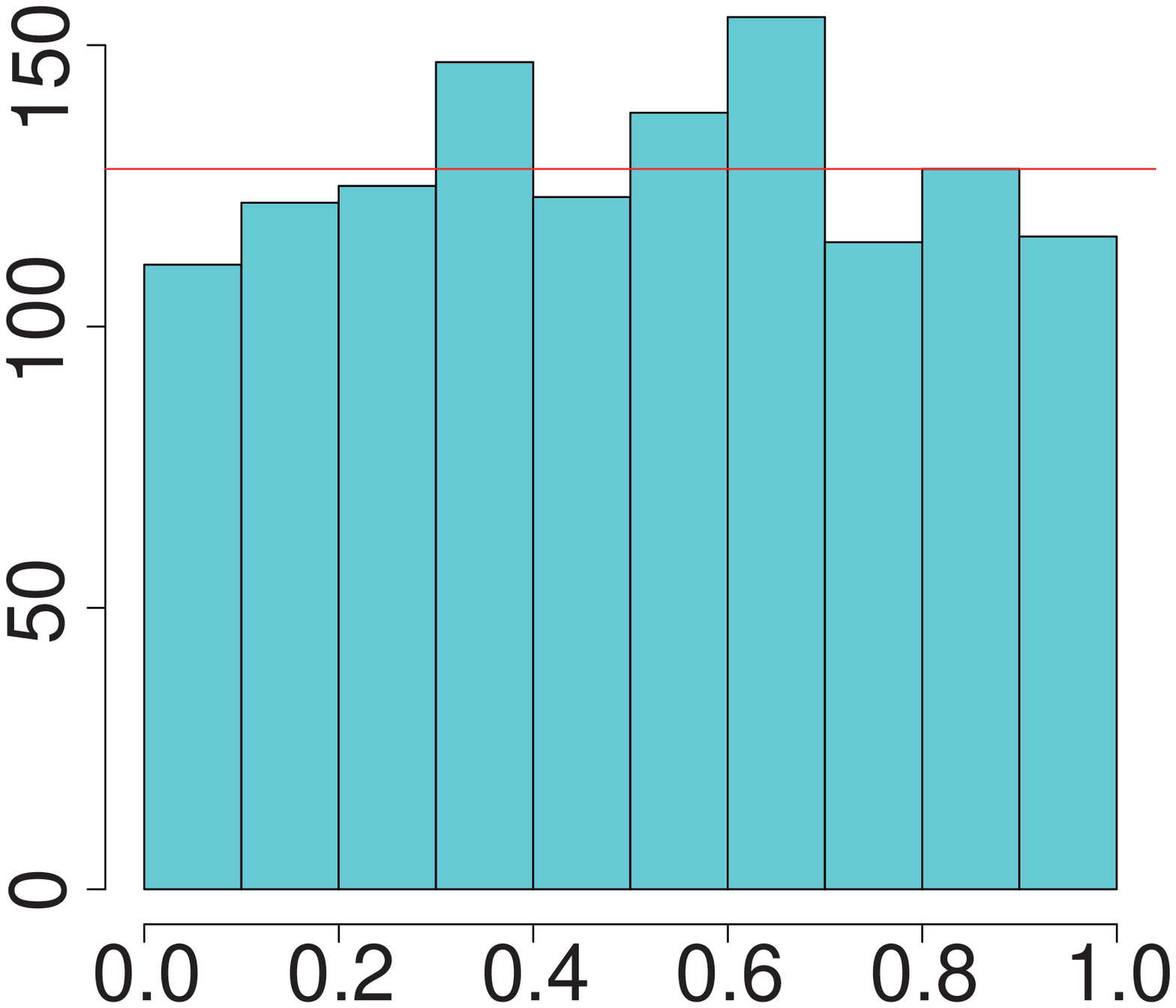}
\end{minipage}

\vspace{5mm}

\begin{minipage}{.25\textwidth}
\centering
MB, $b=0$

\includegraphics[width=1.3in]{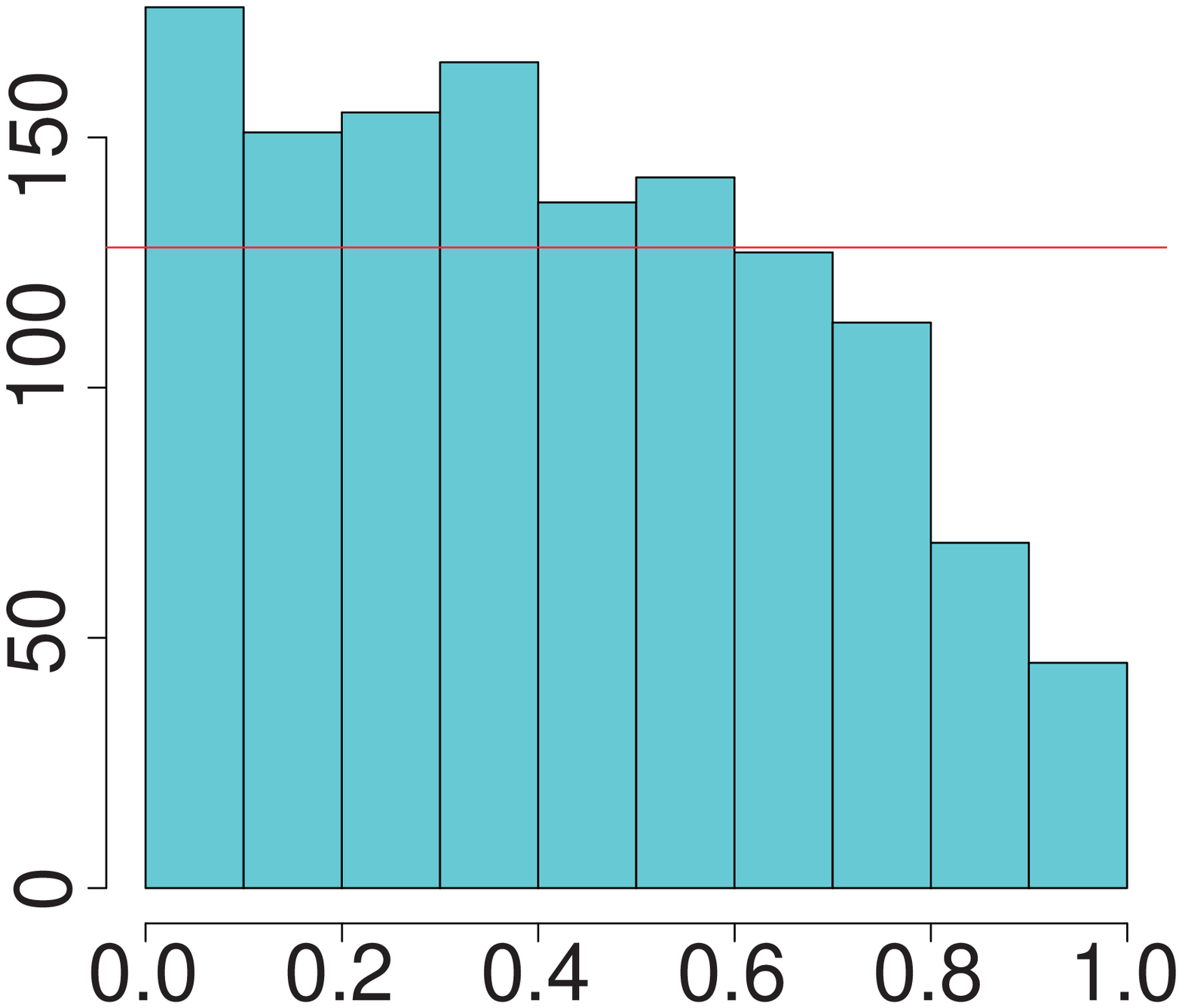}
\end{minipage}
\begin{minipage}{.25\textwidth}
\centering
MB, $b=0.01$

\includegraphics[width=1.3in]{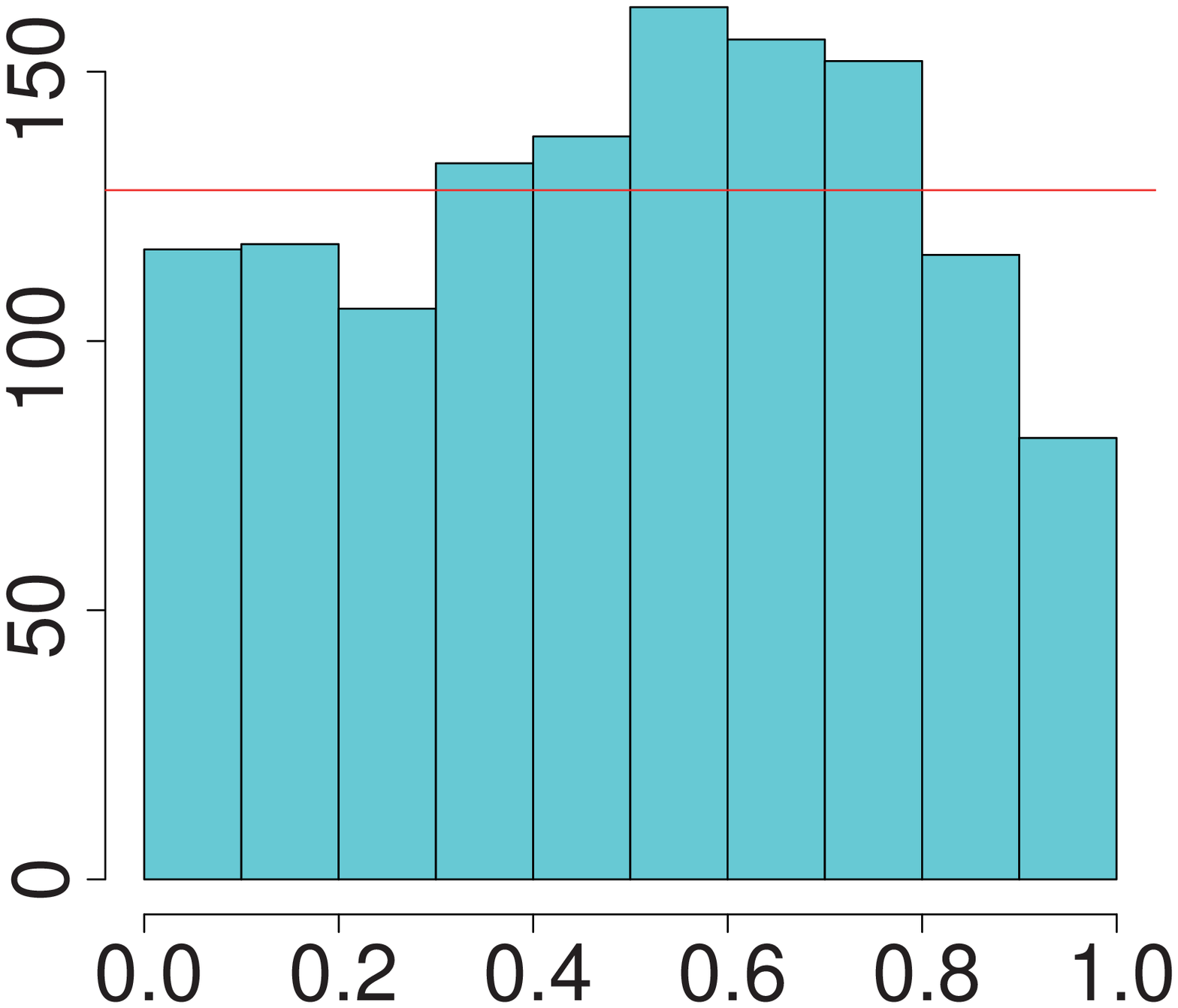}
\end{minipage}
\begin{minipage}{.25\textwidth}
\centering
MB, $b=0.1$

\includegraphics[width=1.3in]{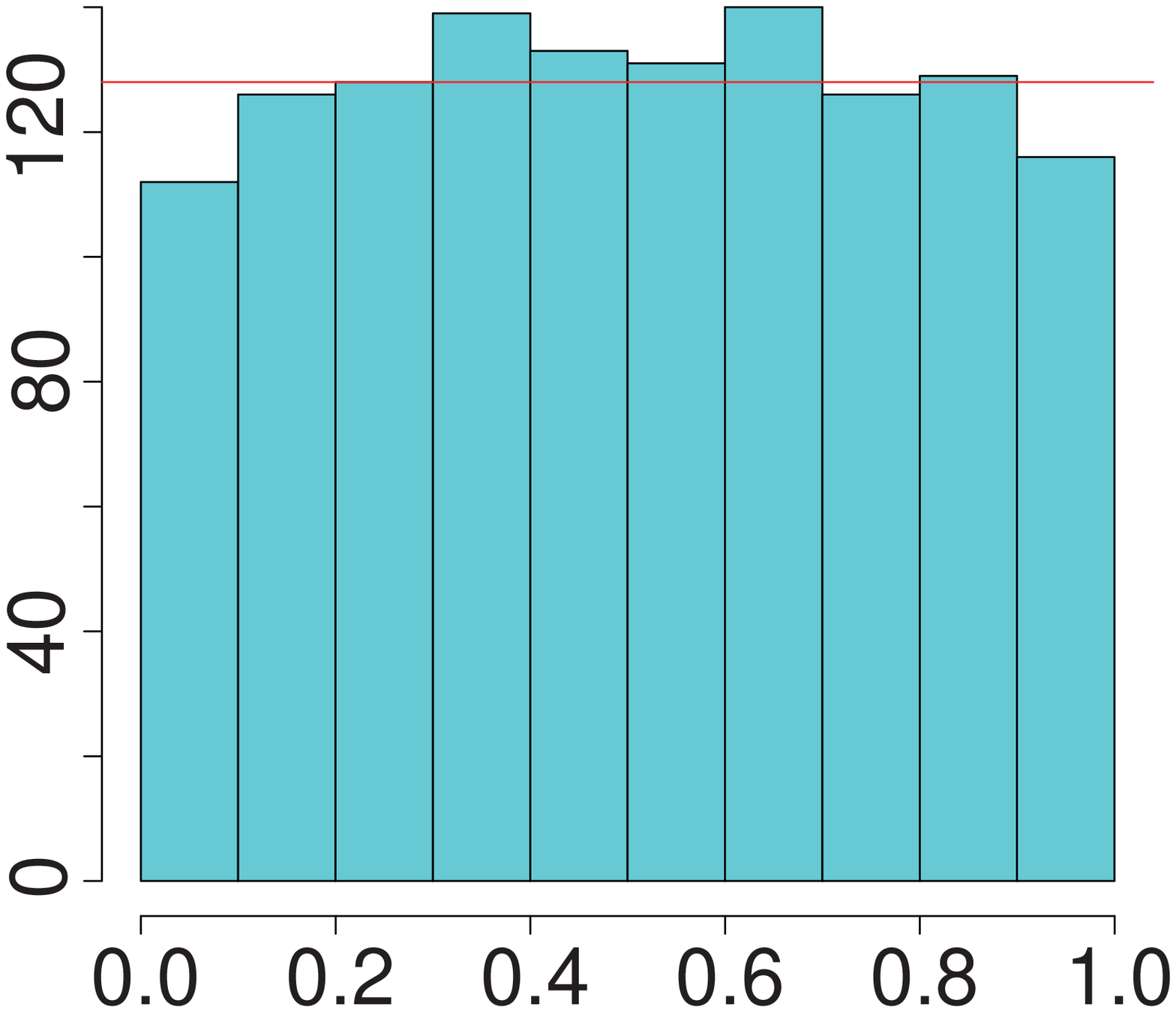}
\end{minipage}
\caption{Top row: Histograms of $p$-values of $H_{21}^{+}$ by ordinary bootstrap (BP). Bottom row:  Histograms of $p$-values of $H_{21}^{+}$ by multiscale bootstrap (MB). The uniform density functions are given by the red lines.}
\label{fig:twovar}
\end{figure}

\begin{figure}[!tb]
\begin{center}
\includegraphics[width=3in]{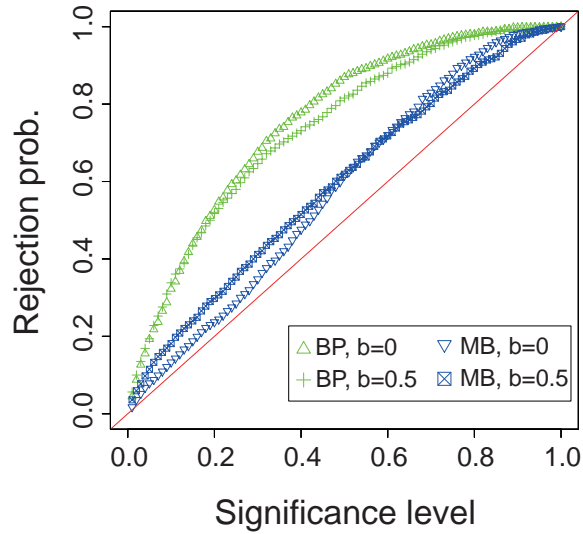}
\caption{Scatterplots of empirical rejection probabilities of $H_{32}^{+}$ by ordinary bootstrap and those by multiscale bootstrap versus significance levels in the six variable cases. 
}
\end{center}
\label{fig:plots}
\end{figure}

The histograms of $p$-values of $H_{21}^{+}$ computed by ordinary bootstrap and those by multiscale bootstrap in the two variable cases are shown in Fig.~\ref{fig:twovar}. Similar histograms were obtained for the other conditions. 
Each of the histograms of $p$-values computed by multiscale bootstrap looked closer to the uniform distribution than by ordinary bootstrap. 
This implied that multiscale bootstrap provided better unbiased $p$-values. 

 In Fig.~3, we also show a scatterplot of empirical rejection probabilities by ordinary bootstrap  {\rm Prob}\{$p^{BP}(H_{32}$)$<$$\alpha$\} and those  by multiscale bootstrap {\rm Prob}\{$p^{MB}_3(H_{32}$)$<$$\alpha$\} versus significance levels $\alpha$ in the six variable cases. 
Plots for unbiased tests should be on the diagonal. 
That is, their rejection probabilities should be equal to the corresponding significance levels.
Most of the plots for ordinary bootstrap are far above the diagonal, indicating that ordinary bootstrap gave rather biased $p$-values and tended to reject reasonable hypotheses much more often than the nominal frequencies or significance levels.
In contrast, the plots for multiscale bootstrap are much closer to the diagonal. This showed that multiscale bootstrap provided much better unbiased $p$-values. 

\section{Conclusion}\label{sec:conc}
We proposed a new procedure to evaluate statistical reliability of LiNGAM. 
Our procedure gives $p$-values of variable orderings estimated by LiNGAM and tells which orderings are more reliable. 
The utility of our procedure was demonstrated in the simulations. 
Future work would investigate how sensitive to non-smoothness of the boundaries of hypothesis regions our method is and how it is alleviated, although the simulations implied that it might be not very problematic. 

\subsubsection*{Acknowledgments}
This research was partially supported by JSPS Global COE program `Computationism as a Foundation for the Sciences'. 

\bibliography{Shohei_Tex_Ref}

\begin{thebibliography}{10}

\bibitem{Bollen89book}
Bollen, K.A.:
\newblock Structural Equations with Latent Variables.
\newblock John Wiley \& Sons (1989)

\bibitem{Pearl00book}
Pearl, J.:
\newblock Causality: Models, Reasoning, and Inference.
\newblock Cambridge University Press (2000)

\bibitem{Spirtes93book}
Spirtes, P., Glymour, C., Scheines, R.:
\newblock Causation, Prediction, and Search.
\newblock Springer Verlag (1993) (2nd ed. MIT Press 2000).

\bibitem{Shimizu06JMLR}
Shimizu, S., Hoyer, P.O., Hyv{\"a}rinen, A., Kerminen, A.:
\newblock A linear non-gaussian acyclic model for causal discovery.
\newblock J. Machine Learning Research \textbf{7} (2006)  2003--2030

\bibitem{Lehmann08Test}
Lehmann, E., Romano, J.:
\newblock Testing Statistical Hypotheses (3rd edition).
\newblock Springer (2008)

\bibitem{Efron93book}
Efron, B., Tibshirani, R.:
\newblock An Introduction to the {Bootstrap}.
\newblock Chapman \& Hall, New York (1993)

\bibitem{Felsenstein85Evol}
Felsenstein, J.:
\newblock Confidence limits on phylogenies: an approach using the bootstrap.
\newblock Evolution \textbf{39} (1985)  783--791

\bibitem{Fried99UAI}
Friedman, N., Goldszmidt, M., Wyner, A.:
\newblock Data analysis with {Bayesian} networks: A bootstrap approach.
\newblock In: Proc. Conf. on Uncertainty in Artificial Intelligence (UAI1999).
  (1999)  196--205

\bibitem{Efron96boot}
Efron, B., Halloran, E., Holmes, S.:
\newblock Bootstrap confidence levels for phylogenetic trees.
\newblock In: Proc. Natl. Acad. Sci. USA. (1996)  13429--13434

\bibitem{Hillis93SB}
Hillis, D., J.Bull:
\newblock An empirical test of bootstrapping as a method for assessing
  confidence in phylogenetic analysis.
\newblock Syst. Biol. \textbf{42} (1993)  182--192

\bibitem{Hall92book}
Hall, P.:
\newblock The bootstrap and Edgeworth expansion.
\newblock Springer-Verlag, New York (1992)

\bibitem{Shimo02SB}
Shimodaira, H.:
\newblock An approximately unbiased test of phylogenetic tree selection.
\newblock Systematic Biology \textbf{51} (2002)  492--508

\bibitem{Shimo08JSPI}
Shimodaira, H.:
\newblock Testing regions with nonsmooth boundaries via multiscale bootstrap.
\newblock J. Statistical Planning and Inference \textbf{138} (2008)  1227--1241

\bibitem{Hoyer07IJAR}
Hoyer, P.O., Shimizu, S., Kerminen, A., Palviainen, M.:
\newblock Estimation of causal effects using linear non-gaussian causal models
  with hidden variables.
\newblock Int. J. Approximate Reasoning \textbf{49} (2008)  362--378

\bibitem{Hyva01book}
Hyv{\"a}rinen, A., Karhunen, J., Oja, E.:
\newblock Independent component analysis.
\newblock Wiley, New York (2001)

\bibitem{Akaike74AIC}
Akaike, H.:
\newblock A new look at the statistical model identification.
\newblock IEEE Trans. Automat. Control \textbf{19} (1974)  716--723

\bibitem{Silva06JMLR}
Silva, R., Scheines, R., Glymour, C., Spirtes, P.:
\newblock Learning the structure of linear latent variable models.
\newblock J. Machine Learning Research \textbf{7} (2006)  191--246

\end{thebibliography}
\bibliographystyle{splncs}



\end{document}